\documentclass{article}

\usepackage{arxiv}

\usepackage[utf8]{inputenc} 
\usepackage[T1]{fontenc}    
\usepackage{hyperref}       
\usepackage{url}            

\usepackage[
  backend=biber,
  style=numeric,
  sorting=none,
  maxbibnames=3,
  minbibnames=1,
  maxcitenames=3,
  mincitenames=1,
  giveninits=true,
  doi=true,
  url=false
]{biblatex}

\DeclareNameAlias{default}{given-family}
\DefineBibliographyStrings{english}{andothers={et\addabbrvspace al\adddot}}
\AtEveryBibitem{\clearfield{urldate}}
\usepackage{booktabs}       
\usepackage{amsfonts}       
\usepackage{nicefrac}       
\usepackage{microtype}      
\usepackage{graphicx}
\usepackage{tikz}
\usetikzlibrary{positioning, arrows.meta, shadows, shapes.geometric, calc}
\usepackage{subcaption}
\usepackage{amsmath}
\usepackage{multirow}
\usepackage{float}
\usepackage{enumitem}
\usepackage{nccmath}
\usepackage{array}
\graphicspath{ {./assets/} }

\title{BPMN Assistant: An LLM-Based Approach to Business Process Modeling}

\author{
 Josip Tomo Licardo \\
  Faculty of Informatics\\
  Juraj Dobrila University of Pula\\
  Zagrebačka 30\\
  52100 Pula, Croatia \\
  \texttt{jlicardo@unipu.hr} \\
   \And
 Nikola Tanković \\
  Faculty of Informatics\\
  Juraj Dobrila University of Pula\\
  Zagrebačka 30\\
  52100 Pula, Croatia \\
  \texttt{ntankov@unipu.hr} \\
  \And
 Darko Etinger \\
  Faculty of Informatics\\
  Juraj Dobrila University of Pula\\
  Zagrebačka 30\\
  52100 Pula, Croatia \\
  \texttt{detinger@unipu.hr} \\
}

\begin{document}
\maketitle
\begin{abstract}
This paper presents BPMN Assistant, a tool that leverages Large Language Models for natural language-based creation and editing of BPMN diagrams. While direct XML generation is common, it is verbose, slow, and prone to syntax errors during complex modifications. We introduce a specialized JSON-based intermediate representation designed to facilitate atomic editing operations through function calling. We evaluate our approach against direct XML manipulation using a suite of state-of-the-art models, including GPT-5.1, Claude 4.5 Sonnet, and DeepSeek V3. Results demonstrate that the JSON-based approach significantly outperforms direct XML in editing tasks, achieving higher or equivalent success rates across all evaluated models. Furthermore, despite requiring more input context, our approach reduces generation latency by approximately 43\% and output token count by over 75\%, offering a more reliable and responsive solution for interactive process modeling.
\end{abstract}


\section{Introduction}

Business Process Model and Notation (BPMN) \cite{BusinessProcessModel} has long served as a standard for modeling business processes, enabling organizations to visualize and optimize their workflows. However, the complexity inherent in creating, editing, and interpreting BPMN diagrams poses significant challenges, particularly for individuals without specialized training \cite{rosemannPotentialPitfallsProcess2006}. As Recker~\cite{reckerOpportunitiesConstraintsCurrent2010} demonstrates through extensive research, BPMN's over-engineered nature and the prevalent lack of formal training among its users create substantial barriers to effective adoption. This bottleneck often results in inefficiencies and a reliance on experts, which can increase operational costs and delay decision-making.

A significant challenge in modern organizations is the communication gap between IT departments and business stakeholders. While IT professionals are comfortable with formal modeling notations and technical specifications, business users typically express processes in natural language and informal descriptions. A systematic review by Njanka et al. \cite{njankaITBusinessAlignmentSystematic2021} demonstrates that this disconnect leads to misunderstandings, requirements misalignment, and implementation delays, with communication barriers persisting despite significant investments in alignment initiatives. Furthermore, valuable process knowledge frequently remains trapped in the minds of domain experts or scattered across various informal documents, making it difficult to capture and formalize this expertise \cite{vanderaaFragmentationProcessInformation2015}.

The complexity of extracting process models from unstructured text remains a significant challenge, as highlighted by Bellan et al.'s qualitative analysis \cite{bellanQualitativeAnalysisState2020}. Their research reveals that current methodologies primarily rely on ad-hoc rule-based approaches, which struggle with the complexity of real-world documents. This challenge is further compounded by the limitations of traditional process elicitation methods, as demonstrated by Baião et al. \cite{fernandaaraujobaiaoLetMeTell}, who propose a novel approach integrating group storytelling techniques with text mining to capture the nuances of human-centric activities in process modeling.

Recent research has highlighted the cognitive challenges faced by modelers when creating process models, particularly regarding cognitive load and task complexity. Weber et al. \cite{weberMeasuringCognitiveLoad2015} introduced a psycho-physiological approach to assess cognitive load during process modeling, utilizing real-time eye movement analysis to identify task-specific difficulties. Their findings emphasize that process modeling requires substantial cognitive effort, especially in naming activities and managing complex structures. These insights reinforce the need for tools like BPMN Assistant, which aim to reduce cognitive barriers and enhance accessibility.

The technical barrier to BPMN usage, when combined with the dispersed nature of process documentation, creates a significant barrier to effective process management. Organizations often find themselves in a situation where valuable process knowledge exists but remains underutilized due to its informal and distributed nature.

Another critical issue is the dynamic nature of business processes in today's rapidly evolving business environment. Organizations need to frequently update and modify their processes to remain competitive and adapt to changing market conditions \cite{desalegnDisentanglingOrganizationalAgility2024}. However, the formal nature of BPMN and the expertise required to modify process models often create a bottleneck in implementing these changes, leading to a gap between actual business operations and their formal documentation. This challenge is evidenced by Leopold et al.'s analysis of 585 BPMN models from industry, which revealed persistent quality issues related to model complexity, particularly in areas such as splits and joins, message flows, and model decomposition, highlighting the technical expertise required for effective BPMN modeling \cite{leopoldLearningQualityIssues2016}.

The disconnect between formal process models and actual business operations is further exacerbated by the increasing complexity of modern business processes. These processes often span multiple departments, involve numerous stakeholders, and integrate with various systems and external partners. As van der Aalst and Weijters \cite{vanderaalstProcessMiningResearch2004} demonstrate in their seminal work on process mining, organizations face significant challenges in accurately capturing and analyzing these complex processes, including issues with hidden tasks, duplicate activities, and non-free-choice constructs. The traditional modeling approaches struggle to address these challenges while maintaining accessibility for all stakeholders.

Prior to the emergence of Large Language Models (LLMs), numerous attempts were made to automate aspects of process modeling and bridge the natural language gap. Early work by Friedrich et al. \cite{friedrichProcessModelGeneration2011} demonstrated the potential of rule-based approaches for extracting process models from natural language text, though these systems struggled with complex sentence structures and domain-specific terminology. The integration of natural language processing into process management faced significant challenges with semantic ambiguity and contextual understanding. Efforts by Leopold et al. \cite{leopoldGeneratingNaturalLanguage2012} to automatically generate natural language descriptions from process models highlighted both the potential and limitations of traditional NLP approaches in process automation. These early automation attempts, while groundbreaking, were often constrained by their reliance on predefined rules and limited ability to handle variations in natural language expressions \cite{mendlingNaturalLanguageProcessing}.

Advancements in artificial intelligence, particularly the emergence of LLMs, have opened new avenues for automating the creation and management of business process models \cite{grohsLargeLanguageModels2024, kampikLargeProcessModels2023, linkonAdvancementsApplicationsGenerative2024}. These models have demonstrated exceptional capabilities in understanding and generating natural language, making them well-suited for bridging the gap between textual process descriptions and formal BPMN representations. Recent research by Klievtsova et al.~\cite{klievtsovaConversationalProcessModeling2023} has shown that AI-assisted approaches to process modeling can be particularly effective, with AI-generated models often being preferred over those created by inexperienced human modelers. However, Rebmann et al.~\cite{rebmannEvaluatingAbilityLLMs2024} highlight that while LLMs show promise in process-related tasks, their effectiveness often depends on proper fine-tuning and task-specific training, particularly for complex semantics-aware operations. This suggests significant potential for LLM-based tools in democratizing access to process modeling capabilities, while also emphasizing the importance of careful implementation and training approaches.

This work presents BPMN Assistant, a system that leverages LLMs to address these challenges. While recent research has demonstrated the potential of AI-assisted process modeling, the critical aspect of model modification and maintenance remains largely unexplored. Our work advances the state of the art by demonstrating the effectiveness of LLMs in editing existing BPMN diagrams through natural language instructions. Rather than framing BPMN modeling as a one-shot generation problem, this work is conceptualized around the idea of incremental process transformation. We argue that BPMN models should be treated as mutable process structures that can be reliably modified through a constrained set of well-defined operations, instead of being repeatedly regenerated in their entirety. This conceptual perspective motivates the separation of process logic from BPMN 2.0 XML syntax and forms the basis for the structured intermediate representation proposed in this study.

\subsection{Research Objectives}

This study addresses the following research questions:
\begin{itemize}
    \item \textbf{RQ1:} How does a structured intermediate representation (JSON) compare to direct standard notation (XML) in terms of generation reliability and editing success rates?
    \item \textbf{RQ2:} To what extent does a function-based editing approach enable open-weights models to perform complex modeling tasks previously reserved for proprietary frontier models?
    \item \textbf{RQ3:} How does the trade-off between increased input context and reduced output complexity impact overall system latency and operational efficiency?
\end{itemize}

\subsection{Contribution}

We propose a novel approach that utilizes a structured, JSON-based intermediate representation to abstract away the syntactic complexity of BPMN 2.0 XML. By shifting the generative task from producing verbose XML to manipulating a concise JSON structure, our system allows LLMs to focus on process logic rather than formatting rules. At a conceptual level, this approach reframes BPMN editing as a controlled transformation problem, where modifications are expressed as atomic logical operations rather than as low-level syntactic changes to XML documents.

Our evaluation reveals significant improvements in editing success rates compared to baseline approaches, particularly when using structured representations for model manipulation. Furthermore, we demonstrate that this approach significantly reduces processing latency and output token costs, making it a viable solution for interactive tools.

The implementation of BPMN Assistant is publicly available at 
\url{https://github.com/jtlicardo/bpmn-assistant}.

The following sections detail our methodology, system implementation, and evaluation results, providing a foundation for future research in AI-assisted process modeling.

\section{Related Work}

The integration of LLMs into Business Process Management (BPM) has led to significant developments in process automation and modeling. The introduction of specialized tools like the BPMN-Chatbot \cite{kopkeIntroducingBPMNChatbotEfficient2024} and ProMoAI \cite{kouraniProMoAIProcessModeling2024} has demonstrated the practical applications of this technology. Initial explorations by Berti and Qafari \cite{bertiLeveragingLargeLanguage2023} into using GPT-4 and Bard (Gemini) for process mining tasks showed promising results in interpreting both procedural and declarative models, highlighting the potential of LLMs in the BPM domain. These developments have led to comprehensive frameworks for process modeling with LLMs \cite{kouraniProcessModelingLarge2024}, showing promising results in both automation and model quality.

\subsection{Process Extraction and Prompting Strategies}

In the area of process extraction, various approaches have demonstrated different methodologies and results. Ferreira et al. \cite{ferreiraSemiautomaticApproachIdentify2017} developed a semi-automatic method for identifying business process elements from natural language texts, achieving 91.92\% accuracy in their prototype implementation through carefully defined mapping rules. Taking a different approach, Neuberger et al. \cite{neubergerUniversalPromptingStrategy2024} introduced a universal prompting strategy that leverages LLMs for process model information extraction, demonstrating performance improvements of up to 8\% F1 score over traditional machine learning methods.

The integration of prompt engineering in BPM has emerged as a promising direction, as discussed by Busch et al. \cite{buschJustTellMe2023}. Their research highlights how prompt engineering can effectively utilize pre-trained language models without extensive fine-tuning, addressing common challenges in process extraction and predictive monitoring. This approach is particularly valuable in scenarios with limited data availability, offering a computationally sustainable solution for BPM applications.

\subsection{Interactive Modeling Tools}

ProMoAI, introduced by Kourani et al. \cite{kouraniProMoAIProcessModeling2024}, represents a significant development in LLM-driven process modeling. The tool prompts an LLM to generate constrained Python code for constructing an intermediate POWL representation and incorporates prompt-engineering and iterative error-handling mechanisms to improve reliability. The resulting models are then transformed into standard notations and can be viewed and exported as BPMN and Petri nets (PNML). 

The framework underpinning ProMoAI was evaluated in a subsequent benchmark study \cite{kouraniEvaluatingLargeLanguage2024}, which compared multiple LLMs using conformance checking (harmonic mean of fitness and precision against ground-truth event logs) as a model-quality measure. In this evaluation, Claude 3.5 Sonnet achieved the highest average quality score (0.93), approaching the ground-truth baseline (0.98). 

Köpke and Safan \cite{kopkeIntroducingBPMNChatbotEfficient2024} introduced the BPMN-Chatbot, a publicly available web-based tool designed for efficient LLM-based process modeling. The BPMN-Chatbot allows users to generate BPMN process models interactively using text or voice input. It achieved notable efficiency gains, reducing token usage by up to 94\% compared to alternative tools like ProMoAI while maintaining a correctness rate of 95\%, which surpassed the best competitor's 86\%. The authors emphasized the importance of user testing to evaluate the feedback loop’s capabilities, which are critical for interactive process design.

While BPMN-Chatbot demonstrates notable strengths, particularly in token efficiency and correctness rates, its empirical evaluation does not isolate the robustness of iterative editing in the feedback loop (e.g., edit success rate over multi-step refinements, structural validity guarantees, or minimal-change behavior), beyond a preliminary technology acceptance test.

Hörner et al. \cite{hornerAutomaticallyGeneratingBPMN2026} introduce BPMNGen, an LLM-based conversational framework for generating BPMN 2.0 process models from natural-language descriptions and iteratively refining them through user prompts. Their work is distinguished by a comprehensive human-centered evaluation, including controlled user studies assessing semantic alignment, cognitive load, acceptability, and comprehension performance of LLM-generated models compared to expert-created ones. The results indicate that LLM-generated models can achieve expert-level semantic quality for simple and moderately complex processes, while expert intervention remains beneficial for highly complex scenarios.

In contrast, BPMN Assistant focuses on a different aspect of the problem: robust and fine-grained interactive BPMN editing, emphasizing structural correctness, editing reliability, latency, and token efficiency across multiple LLM backends. Rather than evaluating model comprehensibility, our work addresses the engineering challenges of incremental model manipulation and validation, which are complementary to the human-centered quality dimensions studied in BPMNGen.

\subsection{Automated Generation and Evaluation Frameworks}

In a comprehensive evaluation of process extraction approaches, Bellan et al. \cite{bellanProcessExtractionText2023} analyzed ten state-of-the-art methods for extracting process models from textual descriptions. Their systematic comparison revealed significant variations in performance and methodology among existing tools, with no single approach achieving superior results across all evaluation metrics. This study highlighted the need for standardized evaluation frameworks and emphasized the challenges in automated process extraction that our work aims to address.

Kourani et al. \cite{kouraniProcessModelingLarge2024} proposed a comprehensive framework leveraging LLMs to automate the generation and refinement of process models from textual descriptions. This framework demonstrated its ability to streamline process modeling tasks while maintaining sound and executable model outputs. The superiority of this approach was evident in its comparison with traditional methods, particularly in resolving errors and integrating user feedback effectively. GPT-4 showcased strong performance in generating process models, addressing errors, and adapting to user feedback, whereas Gemini struggled with similar tasks.

Nivon et al. \cite{nivonAutomatedGenerationBPMN2025} introduced a novel approach to automating BPMN process generation from textual requirements, addressing the challenges faced by non-expert users in process modeling. Their methodology employs a three-step approach utilizing a fine-tuned GPT-3.5 model: first extracting tasks and ordering constraints from textual descriptions, then constructing an abstract syntax tree (AST) to represent task relationships, and finally converting the AST into a BPMN process. Their Java-based implementation achieved a 78.5\% accuracy rate for valid BPMN processes in tests across 200 descriptions, demonstrating the viability of automated BPMN generation while highlighting areas for potential improvement through more advanced language models.

These studies collectively illustrate the transformative potential of LLMs in business process modeling. Tools like BPMN-Chatbot, ProMoAI, and Nivon et al.'s automated BPMN generator demonstrate the feasibility of combining generative AI with domain-specific methodologies to achieve higher efficiency, accuracy, and accessibility. Moreover, the benchmarks and frameworks established by researchers provide a foundation for future advancements, emphasizing the importance of careful prompt design, user feedback integration, and domain-specific optimization in realizing the full potential of LLMs in BPM.

\begin{table}[h!]
\centering
\small
\renewcommand{\arraystretch}{1.6}
\begin{tabular}{@{}>{\raggedright\arraybackslash}p{2.8cm} >{\raggedright\arraybackslash}p{3.2cm} >{\raggedright\arraybackslash}p{3.2cm} >{\raggedright\arraybackslash}p{3.2cm} >{\raggedright\arraybackslash}p{3.2cm}@{}}
\toprule
\textbf{Axis} & \textbf{ProMoAI \cite{kouraniProMoAIProcessModeling2024, kouraniEvaluatingLargeLanguage2024}} & \textbf{BPMN-Chatbot \cite{kopkeIntroducingBPMNChatbotEfficient2024}} & \textbf{BPMNGen \cite{hornerAutomaticallyGeneratingBPMN2026}} & \textbf{BPMN Assistant} \\
\midrule
Primary goal
& NL$\rightarrow$model generation + refinement/optimization
& efficient NL$\rightarrow$BPMN
& NL$\rightarrow$BPMN + human-centered quality
& reliable NL$\rightarrow$BPMN + \textbf{edit robustness} \\
Output notation
& BPMN + PNML
& BPMN
& BPMN
& BPMN \\
Evaluation focus
& conformance-based model quality
& correctness + token efficiency + acceptance
& semantic alignment + cognitive load + acceptability + comprehension
& GED/RGED structural fidelity + failure rate \\
Evaluation method
& automated
& automated + user study
& user study
& automated \\
\bottomrule
\end{tabular}
\vspace{0.2cm}
\caption{Comparison of interactive modeling tools by research objective and evaluation emphasis.}
\label{tab:comparison_focus}
\end{table}

\section{System Architecture}

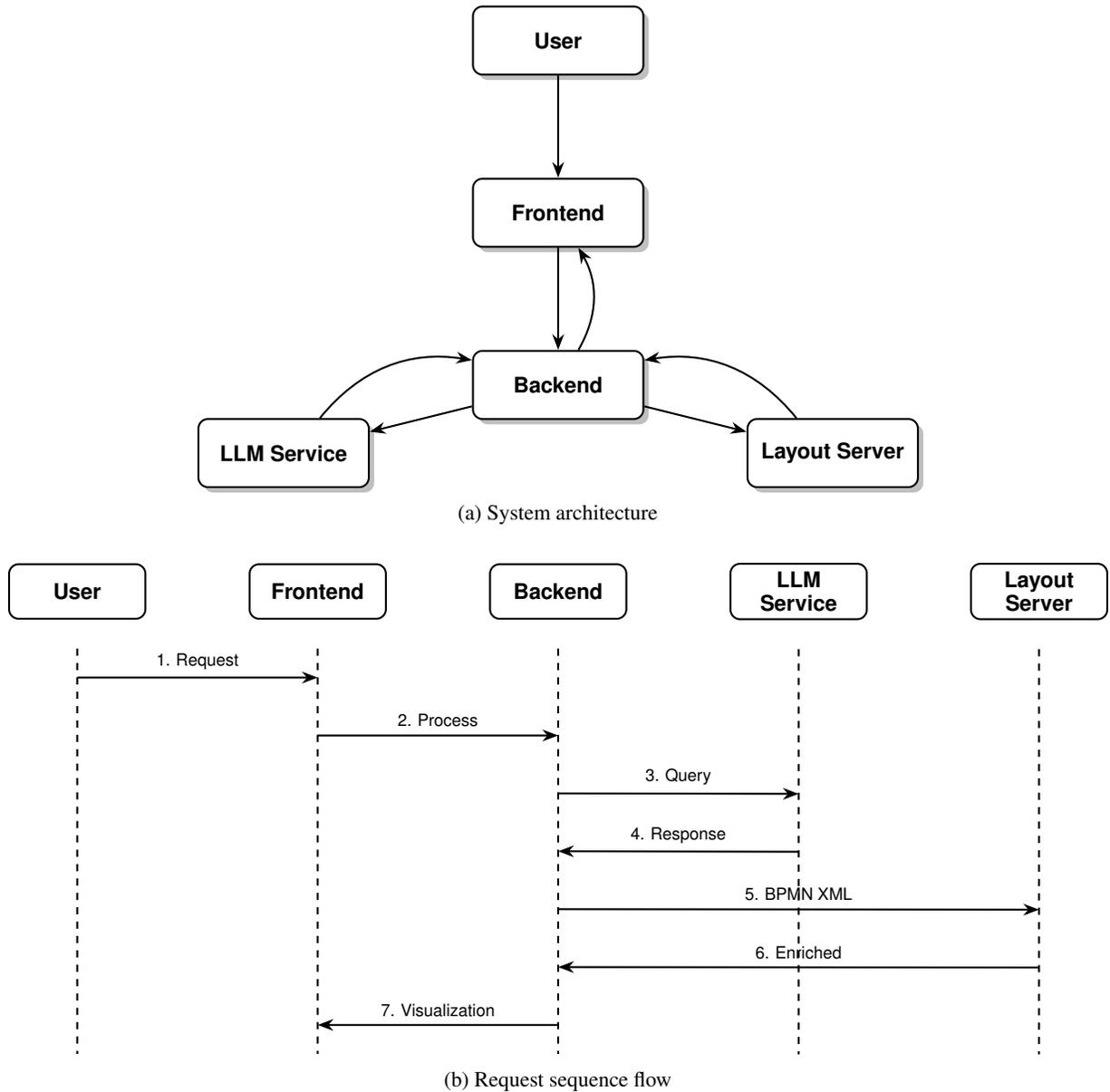
\begin{figure}[htbp]
\centering
\begin{subfigure}{\textwidth}
\centering
\begin{tikzpicture}[
    node distance=1.5cm,
    comp/.style={
        rectangle,
        draw,
        thick,
        rounded corners,
        drop shadow,
        fill=white,
        minimum height=1cm,
        minimum width=2.5cm,
        align=center,
        font=\sffamily\bfseries\small
    },
    conn/.style={
        ->,
        >=Stealth,
        thick
    },
    scale=0.9
]
\node[comp] (user) {User};
\node[comp, below=of user] (frontend) {Frontend};
\node[comp, below=of frontend] (backend) {Backend};
\node[comp, left=of backend, yshift=-1cm] (llm) {LLM Service};
\node[comp, right=of backend, yshift=-1cm] (layout) {Layout Server};

\draw[conn] (user) -- (frontend);
\draw[conn] (frontend) -- (backend);
\draw[conn] (backend) to[bend right] (frontend);
\draw[conn] (backend) -- (llm);
\draw[conn] (backend) -- (layout);
\draw[conn] (llm) to[bend left] (backend);
\draw[conn] (layout) to[bend right] (backend);
\end{tikzpicture}
\caption{System architecture}
\label{fig:simple-arch}
\end{subfigure}

\vspace{0.5cm}
\begin{subfigure}{\textwidth}
\centering
\begin{tikzpicture}[
    node distance=1.5cm,  
    box/.style={
        rectangle,
        draw,
        thick,
        rounded corners,
        fill=white,
        minimum height=0.8cm,
        minimum width=2cm,  
        align=center,
        font=\sffamily\bfseries\footnotesize  
    },
    lifeline/.style={
        dashed,
        thick
    },
    message/.style={
        ->,
        >=Stealth,
        thick
    },
    scale=0.85  
]
\node[box] (user) {User};
\node[box, right=of user] (frontend) {Frontend};
\node[box, right=of frontend] (backend) {Backend};
\node[box, right=of backend] (llm) {LLM\\Service};
\node[box, right=of llm] (layout) {Layout\\Server};

\coordinate (user-start) at ($(user.south)+(0,-0.5)$);
\coordinate (user-end) at ($(user-start)+(0,-7)$);
\draw[lifeline] (user-start) -- (user-end);

\coordinate (frontend-start) at ($(frontend.south)+(0,-0.5)$);
\coordinate (frontend-end) at ($(frontend-start)+(0,-7)$);
\draw[lifeline] (frontend-start) -- (frontend-end);

\coordinate (backend-start) at ($(backend.south)+(0,-0.5)$);
\coordinate (backend-end) at ($(backend-start)+(0,-7)$);
\draw[lifeline] (backend-start) -- (backend-end);

\coordinate (llm-start) at ($(llm.south)+(0,-0.5)$);
\coordinate (llm-end) at ($(llm-start)+(0,-7)$);
\draw[lifeline] (llm-start) -- (llm-end);

\coordinate (layout-start) at ($(layout.south)+(0,-0.5)$);
\coordinate (layout-end) at ($(layout-start)+(0,-7)$);
\draw[lifeline] (layout-start) -- (layout-end);

\draw[message] ($(user-start)+(0,-0.5)$) -- ($(frontend-start)+(0,-0.5)$) 
    node[midway, above, font=\sffamily\scriptsize] {1. Request};
\draw[message] ($(frontend-start)+(0,-1.5)$) -- ($(backend-start)+(0,-1.5)$) 
    node[midway, above, font=\sffamily\scriptsize] {2. Process};
\draw[message] ($(backend-start)+(0,-2.5)$) -- ($(llm-start)+(0,-2.5)$) 
    node[midway, above, font=\sffamily\scriptsize] {3. Query};
\draw[message] ($(llm-start)+(0,-3.5)$) -- ($(backend-start)+(0,-3.5)$) 
    node[midway, above, font=\sffamily\scriptsize] {4. Response};
\draw[message] ($(backend-start)+(0,-4.5)$) -- ($(layout-start)+(0,-4.5)$) 
    node[midway, above, font=\sffamily\scriptsize] {5. BPMN XML};
\draw[message] ($(layout-start)+(0,-5.5)$) -- ($(backend-start)+(0,-5.5)$) 
    node[midway, above, font=\sffamily\scriptsize] {6. Enriched};
\draw[message] ($(backend-start)+(0,-6.5)$) -- ($(frontend-start)+(0,-6.5)$) 
    node[midway, above, font=\sffamily\scriptsize] {7. Visualization};
\end{tikzpicture}
\caption{Request sequence flow}
\label{fig:sequence}
\end{subfigure}

\caption{BPMN Assistant system: (a) Component architecture showing key system elements and (b) request sequence flow showing the temporal order of interactions.}
\label{fig:system_arch}
\end{figure}

The system architecture is designed to operationalize the core conceptual assumption of this work: that reliable BPMN modeling with LLMs requires explicit separation between process semantics, editing logic, and concrete BPMN serialization. The BPMN Assistant system architecture, as illustrated in Figure~\ref{fig:system_arch}, is designed to facilitate seamless interaction between users and BPMN models through natural language inputs. The system is composed of three primary components: the Python-based backend, the BPMN layout server, and the Vue.js frontend. Each component plays a critical role in ensuring the system’s functionality, efficiency, and usability.

\subsection{Backend}

The backend, implemented in Python, serves as the core computational engine of the system. Its primary responsibilities include handling user inputs, interacting with the LLM, managing BPMN diagrams, and providing APIs for the frontend. The backend leverages the FastAPI framework, chosen for its performance and simplicity in developing RESTful APIs.

The backend serves as the first point of contact for user inputs, analyzing these inputs to determine the user’s intent. This can include conversational queries, such as asking questions about BPMN concepts, or operational commands, such as creating or modifying BPMN diagrams. When a conversational intent is recognized, the backend communicates with the LLM to generate a natural language response. If the input requires an operational response, the backend constructs a prompt to instruct the LLM to generate or modify a BPMN diagram. These diagrams are initially represented in JSON format and subsequently converted to BPMN XML for compatibility with visualization tools. Additionally, the backend supports the uploading of BPMN files that conform to the supported BPMN subset, enabling users to query or modify existing diagrams within the constraints of the system’s JSON representation. The backend includes robust error handling and validation mechanisms to ensure that outputs from the LLM are accurate and reliable. If invalid JSON is generated, the backend retries the process or notifies the user of the error. By integrating with the BPMN layout server, the backend also ensures that diagrams are enriched with graphical information for enhanced visualization.

\paragraph{Model Integration and Selection}

Our system supports the usage of a wide range of LLMs from different providers, each chosen for its specific strengths and use cases. Table~\ref{tab:model_overview} provides an overview of the supported models and their capabilities.

\begin{table}[h!]
    \centering
    \begin{tabular}{@{}lll@{}}
        \toprule
        \textbf{Provider} & \textbf{Model} & \textbf{Description} \\
        \midrule
        \multirow{3}{*}{OpenAI} 
        & GPT-5.1 & OpenAI's flagship model for coding and agentic tasks. \\
        & GPT-5 mini & Faster, more cost-efficient version of GPT-5, used for well-defined tasks. \\
        & GPT-4o & Earlier-generation multimodal model with broad cross-modal capabilities \cite{openaiGPT4oSystemCard2024}. \\
        \addlinespace
        \multirow{1}{*}{Anthropic} 
        & Claude 3.5 Sonnet & Earlier-generation model with strong reasoning and long-context capabilities. \\
        & Claude 4.5 Sonnet & Frontier model optimized for agentic reasoning and complex coding tasks. \\
        \addlinespace
        \multirow{1}{*}{Google} 
        & Gemini 2.0 Flash & Multimodal model with with low latency and a 1M context window. \cite{comaniciGemini25Pushing2025} \\
        \addlinespace
        \multirow{4}{*}{Fireworks AI} 
        & Llama 3.3 70B & Meta’s open-source text-only LLM \cite{grattafioriLlama3Herd2024}. \\
        & Qwen 2.5 72B & Alibaba’s open-source model with strong multilingual capabilities \cite{qwenQwen25TechnicalReport2025}. \\
        & DeepSeek V3 & Open-source model specialized in technical and scientific reasoning \cite{deepseek-aiDeepSeekV3TechnicalReport2024}. \\
        \bottomrule
    \end{tabular}
    \vspace{0.2cm}
    \caption{Overview of Integrated AI Models}
    \label{tab:model_overview}
\end{table}

\paragraph{Supported BPMN Elements}

The backend facilitates operations on a wide range of BPMN elements to support diverse modeling scenarios. In the category of tasks, the system supports generic tasks, user tasks, service tasks, send tasks, receive tasks, business rule tasks, manual tasks, and script tasks. To manage process flow logic, the system supports exclusive, parallel, and inclusive gateways.

The system also provides support for events, distinguishing between start, end, and intermediate types. Specifically, it supports generic, timer, and message start events; generic and message end events; as well as generic and message intermediate throw events, and generic, timer, and message intermediate catch events. This selection of elements enables the modeling of specific executable processes and event-driven workflows.

\subsection{BPMN Layout Server}

The BPMN layout server, implemented using Node.js with the Express.js framework, is dedicated to augmenting BPMN diagrams with graphical information. This server employs the \texttt{bpmn-auto-layout}\footnote{\url{https://github.com/bpmn-io/bpmn-auto-layout}} npm library to generate DI (Diagram Interchange) information, which includes the graphical coordinates of BPMN elements.

The layout server takes BPMN XML files from the backend and enhances them by adding graphical coordinates, which are essential for visual representation. Although the server efficiently enriches diagrams for most scenarios, it is currently limited in its ability to process multi-pool or multi-lane diagrams due to constraints in the \texttt{bpmn-auto-layout} library. The server is designed to work seamlessly with the backend, exchanging data through REST APIs to ensure smooth integration.

\subsection{Frontend}

The Vue.js-based frontend provides an intuitive graphical user interface (GUI) for user interaction. Its design focuses on accessibility, enabling non-experts to engage with BPMN modeling tasks effectively.

\begin{figure}[h!]
    \centering
    \includegraphics[width=\textwidth]{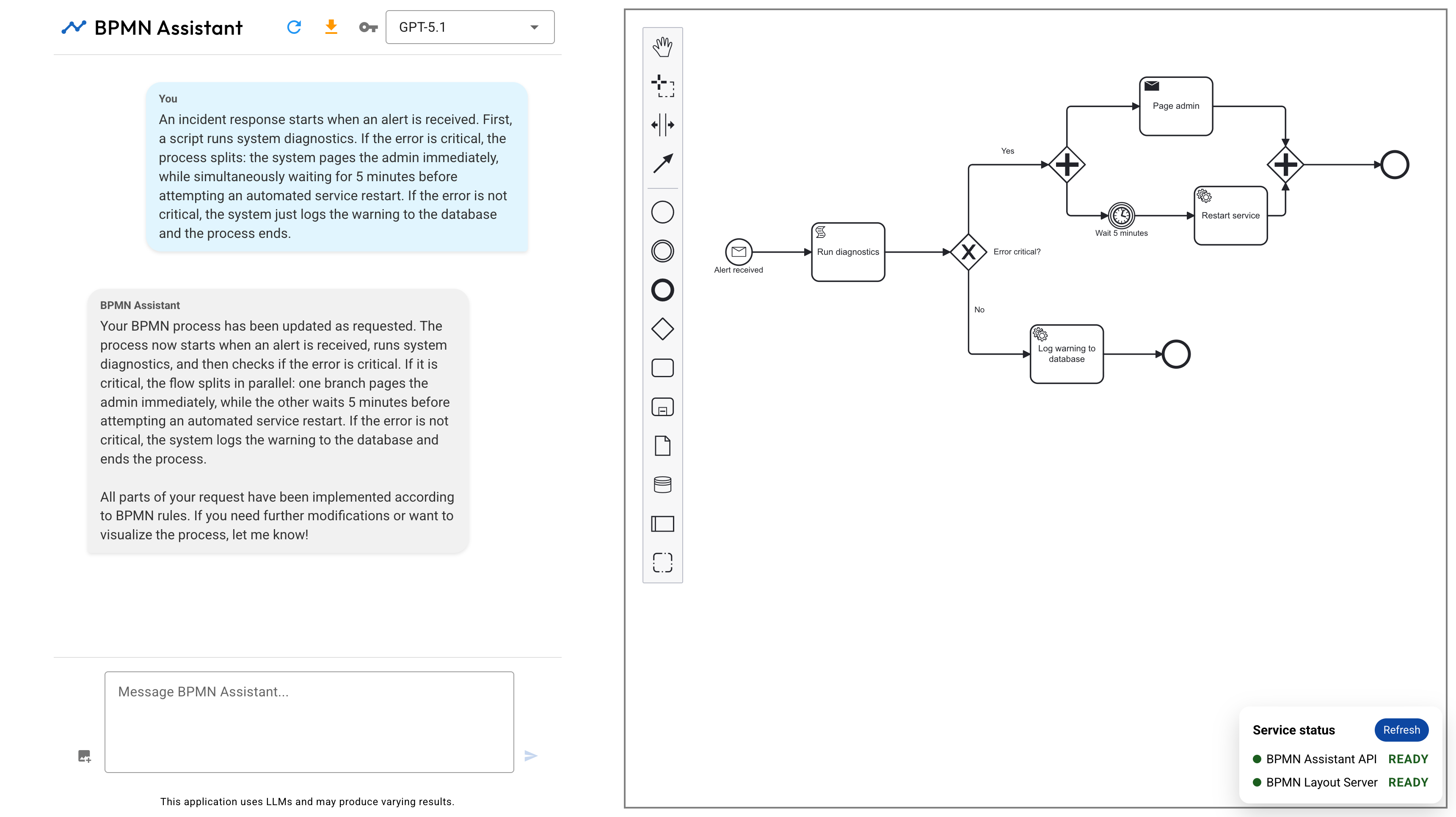}
    \caption{The web application interface featuring a dual-panel design: a chat interface on the left and a BPMN canvas on the right.}
    \label{fig:web_app_interface}
\end{figure}

The frontend features a dual-panel design that facilitates interaction with both the LLM and BPMN diagrams. On the left, a chat interface allows users to submit natural language queries and view responses in real-time. Users can also select their preferred LLM for processing queries. On the right, a BPMN canvas, powered by \texttt{bpmn.io} (bpmn-js), displays the generated or modified BPMN diagrams. This canvas enables users to interact with the diagrams in a manner similar to traditional desktop tools. The frontend also provides feedback to users during diagram generation or modification, displaying status messages that indicate progress. Once a BPMN diagram is complete, users can download it for offline use or integration with other tools.

The frontend bridges the gap between users and the underlying system, providing a user-friendly interface for BPMN modeling and interaction.

\subsection{Data Flow}

The system processes data through a structured sequence of interactions, as illustrated in Figure~\ref{fig:system_arch}. The process initiates when a user submits a natural language request (Step 1). The frontend forwards this to the backend (Step 2), which performs intent recognition. If a modeling task is identified, the backend queries the LLM Service (Step 3). Unlike standard chatbots, the LLM is instructed to return structured JSON data representing the process or specific editing commands (Step 4). The backend then converts this JSON into standard BPMN 2.0 XML and transmits it to the Layout Server (Step 5). The layout server calculates the X/Y coordinates for the diagram elements and returns the enriched XML (Step 6). Finally, the backend delivers the renderable XML to the frontend for visualization (Step 7).

\subsection{JSON Representation of BPMN Diagrams}

The JSON intermediate representation is not merely an implementation convenience, but a conceptual abstraction that captures BPMN control-flow semantics independently of any concrete serialization format. BPMN Assistant utilizes a hierarchical JSON representation to abstract the verbose XML syntax of standard BPMN 2.0. This structure is designed to be easily generated by LLMs while maintaining sufficient fidelity to map back to valid XML. The representation uses a sequence of elements to describe the process flow. Unless branching logic is introduced via gateways, elements in the \texttt{process} array are executed sequentially. By elevating process structure to a first-class representation, this abstraction allows LLMs to reason about process logic directly, without being exposed to the syntactic complexity of BPMN XML.

\paragraph{Tasks}

Tasks represent atomic units of work. The system supports a comprehensive set of task types to handle various modeling scenarios. Beyond the generic \texttt{task}, the system supports: \texttt{userTask}, \texttt{serviceTask}, \texttt{sendTask}, \texttt{receiveTask}, \texttt{businessRuleTask}, \texttt{manualTask}, and \texttt{scriptTask}.

\begin{verbatim}
{
    "type": "userTask", // or serviceTask, sendTask, etc.
    "id": "task_123",
    "label": "Approve request"
}
\end{verbatim}

\paragraph{Events}

The system distinguishes between start, end, and intermediate events. Crucially, it supports \texttt{eventDefinition} attributes to define specific triggers, such as timers or messages. This allows for the modeling of event-driven architectures.

\begin{verbatim}
{
    "type": "intermediateCatchEvent",
    "id": "event_timer",
    "label": "Wait 24 hours",
    "eventDefinition": "timerEventDefinition"
}
\end{verbatim}

\paragraph{Gateways}

Gateways manage flow divergence and convergence. The system supports exclusive (XOR), inclusive (OR), and parallel gateways.

A unique feature of our representation is the \texttt{has\_join} boolean attribute. Since the JSON structure is hierarchical (nested), \texttt{has\_join:true} explicitly signals the XML transformer that the branches merge back into a single flow after execution, triggering the generation of a converging gateway node. If \texttt{false}, the branches may end independently or loop back to previous elements.

To support cyclic flows (loops) within a nested structure, branches utilize an optional \texttt{next} field. This field contains the ID of a target element, allowing the flow to jump to any existing node in the process, breaking the strict hierarchy when necessary.

\begin{verbatim}
{
    "type": "exclusiveGateway", // or "inclusiveGateway"
    "id": "gateway_1",
    "label": "Is approved?",
    "has_join": true,
    "branches": [
        {
            "condition": "Yes",
            "path": [ ... ] // Nested sequence of elements
        },
        {
            "condition": "No",
            "path": [], 
            "next": "task_start" // Example of a loop-back
        }
    ]
}
\end{verbatim}

For inclusive gateways, an \texttt{is\_default} boolean field is available to designate the default flow path.

Parallel gateways are represented as an array of arrays, where each sub-array constitutes a concurrent execution path. Synchronization is handled implicitly: if the parent flow continues after the gateway, a converging parallel gateway is automatically generated in the XML output.

\begin{verbatim}
{
    "type": "parallelGateway",
    "id": "parallel_1",
    "branches": [
        [ { "type": "task", "label": "Path A" } ],
        [ { "type": "task", "label": "Path B" } ]
    ]
}
\end{verbatim}

This structured JSON representation serves as the backbone of the BPMN Assistant, allowing for the accurate and efficient translation of user inputs into process models.

For a more complete example of the BPMN JSON representation, please refer to the Appendix (Section~\ref{sec:appendix-process-example}).

\subsection{Process Editing Functions}

The system supports a set of specialized functions for modifying BPMN diagrams. These functions, outlined in Table~\ref{tab:edit_functions}, enable precise control over process elements while maintaining the structural integrity of the diagram. From a conceptual standpoint, these editing functions define the minimal set of operations required to express meaningful BPMN model transformations while preserving process soundness. When the LLM receives an editing request, it analyzes the natural language input and determines which function(s) to call to achieve the desired modifications.

\begin{table}[h!]
\centering
\begin{tabular}{@{}lll@{}}
\toprule
\textbf{Function} & \textbf{Parameters} & \textbf{Description} \\
\midrule
delete\_element & element\_id & Removes a specified element from the process \\
\addlinespace
redirect\_branch & branch\_condition, next\_id & Redirects a gateway branch to a new target element \\
\addlinespace
add\_element & element, before\_id*, after\_id* & Adds a new element at a specified position \\
\addlinespace
move\_element & element\_id, before\_id*, after\_id* & Relocates an existing element within the process \\
\addlinespace
update\_element & new\_element & Updates properties of an existing element \\
\bottomrule
\multicolumn{3}{l}{\small *Optional parameters, only one should be provided} \\
\end{tabular}
\vspace{0.2cm}
\caption{Process Editing Functions}
\label{tab:edit_functions}
\end{table}

The LLM processes editing requests through a structured approach by first analyzing the user's natural language request to understand the desired changes. It then identifies the affected elements in the current process and determines the appropriate editing functions required to achieve the requested modification. Once identified, the model generates the function calls with the correct parameters and ensures that the proposed changes maintain process integrity through validation.

Each editing function is designed to perform a specific type of modification while preserving the process's logical flow. For example, when deleting an element, the system automatically handles the reconnection of surrounding elements to maintain process continuity. Similarly, when adding new elements, the system ensures proper integration with existing process flows.

The granular nature of these functions allows the LLM to decompose complex editing requests into a series of atomic operations. This approach enhances reliability and makes it easier to validate and verify changes before they are applied to the process model.

\subsection{Validation and Soundness}

To ensure the generation of syntactically sound BPMN models, the system incorporates a strict validation layer implemented in Python that intercepts the LLM output prior to XML conversion. This validator enforces core BPMN structural constraints, including the uniqueness of element identifiers, the correctness of connectivity between flow elements, the validity of gateway branch hierarchies, and the requirement that each process contains exactly one start event. When a violation is detected, the system initiates a self-correction loop in which the validation error is fed back to the LLM, prompting it to revise the intermediate representation. By acting as a programmatic guardrail, this validation mechanism ensures that the final BPMN XML is both syntactically valid and reliably interpretable by standard BPMN engines.

\section{Evaluation}

The evaluation of BPMN Assistant focuses on assessing its accuracy through Graph Edit Distance (GED) and Relative Graph Edit Distance (RGED), two widely recognized metrics for measuring process model similarity \cite{licardoMethodExtractingBPMN2024, dijkmanGraphMatchingAlgorithms2009, sohailIntelligentGraphEdit2021}. Traditional GED methods often struggle with gateway semantics and execution probabilities, limiting their effectiveness in BPMN similarity measurements \cite{waspadaImprovedMethodGraph2020}. Schoknecht et al. \cite{schoknechtSimilarityBusinessProcess2018} provide a comprehensive review of process model similarity techniques, emphasizing that hybrid approaches—integrating syntactic, semantic, and behavioral comparisons—tend to produce more reliable results. However, many of these methods require domain-specific adaptations to be effective for BPMN.

Our evaluation methodology deliberately focuses on structural and syntactic assessment through graph similarity measures. This approach was selected based on the research objectives of comparing intermediate representations for BPMN generation and modification. It is important to note that this evaluation does not encompass semantic evaluation approaches, such as those based on Petri nets or conformance checking from process mining. Similarly, we do not focus on execution performance through simulation studies, nor do we conduct comprehensive usability evaluations. While these aspects represent valuable dimensions for assessing process modeling tools, they fall outside the scope of the current research, which primarily examines the efficacy of different representation approaches in generating structurally accurate BPMN diagrams from natural language descriptions. The selected evaluation metrics allow for direct comparison between JSON and XML approaches while providing insight into the structural correctness of the generated models.

\subsection{Dataset Composition}
To ensure a robust evaluation across diverse scenarios, we constructed a dataset of 60 process descriptions spanning 20 distinct business domains (e.g., Marketing, Healthcare, Logistics). The dataset was curated to evaluate the system's ability to handle fundamental BPMN control-flow constructs, including sequential tasks, exclusive decision points (XOR-split/join), and concurrent execution paths (AND-split/join). This focus on core structural patterns provides a baseline for assessing the generation reliability of the underlying JSON representation.

The evaluation set was generated using a fixed prompt template and the OpenAI \texttt{gpt-4.1} model. Each description was constrained to 7--8 activities and formulated in natural language while explicitly discouraging BPMN-specific terminology. Three descriptions were generated for each of the 20 business domains, after which the outputs were manually screened to remove ambiguous or degenerate cases.

Ground-truth BPMN diagrams were then created from these descriptions by multiple BPMN-trained annotators, including the authors, following a consistent modeling guideline. All diagrams were subsequently reviewed and corrected by the authors to ensure syntactic validity and semantic alignment with the corresponding textual descriptions.

\subsection{Graph Edit Distance (GED) and Relative Graph Edit Distance (RGED)}
GED quantifies the cost of transforming one BPMN diagram into another by counting the minimum number of graph edit operations (node insertion, deletion, or substitution) required. To address the semantic variability inherent in natural language generation (e.g., an LLM generating "Process Order" versus "Handle Order"), we implemented a custom two-stage evaluation pipeline\footnote{Implementation available at: \url{https://github.com/jtlicardo/bpmn-ged}}.

First, we perform semantic label normalization. Raw BPMN XML files are parsed into a graph structure, and a lightweight LLM (GPT-5 mini) is employed to map semantically equivalent labels to identical abstract tokens (e.g., mapping both "Submit Order" and "Send Order" to token "A"). This ensures that the metric evaluates logical flow accuracy rather than penalizing surface-level lexical differences.

Second, we compute the GED using the NetworkX library with a specific cost configuration designed to distinguish between minor syntactic errors and major structural hallucinations. The cost functions are defined as follows:

\begin{description}[leftmargin=4.2cm, style=multiline, font=\bfseries]
    
    \item[Insertion and Deletion] 
    The cost for inserting or deleting a node or edge is set to $1.0$.
    
    \item[Node Substitution] 
    The substitution cost $c_{sub}(n_1, n_2)$ between two nodes is determined by:
    \begin{fleqn}[1em] 
    \begin{equation}
        c_{sub}(n_1, n_2) = 
        \begin{cases} 
        0.0 & \text{if } \text{label}(n_1) = \text{label}(n_2) \land \text{type}(n_1) = \text{type}(n_2) \\
        0.5 & \text{if } \text{label}(n_1) = \text{label}(n_2) \land \text{type}(n_1) \neq \text{type}(n_2) \\
        1.0 & \text{otherwise}
        \end{cases}
    \end{equation}
    \end{fleqn}
    
    \item[Edge Matching] 
    Edges are matched based on the equality of their normalized labels.

\end{description}

The partial penalty of $0.5$ for type mismatches allows us to distinguish cases where the model correctly inferred the intent (the label) but selected the wrong BPMN element type (e.g., using a generic \textit{Task} instead of a \textit{User Task}) from cases where the model hallucinated an entirely incorrect step.

RGED normalizes the GED by considering the complexity of the involved graphs, providing a metric that is independent of diagram size:

\begin{equation}
\text{Relative GED} = \frac{\text{GED}(G_1, G_2)}{\text{GED}(G_1, \emptyset) + \text{GED}(G_2, \emptyset)}
\label{eq:rged}
\end{equation}

\begin{equation}
\text{Similarity} = 1 - \text{Relative GED}
\label{eq:similarity}
\end{equation}

where $\emptyset$ represents an empty graph. This results in a similarity score where $1.0$ represents a semantically identical structure and $0.0$ represents total dissimilarity.

To implement the calculation, we utilized the NetworkX Python library, which provides efficient algorithms for computing graph edit distance. Unlike prior approaches that rely on domain-specific heuristics or ML-based enhancements \cite{sohailIntelligentGraphEdit2021}, our implementation applies standard GED operations to measure similarity between BPMN diagrams, albeit with the pre-processing step of semantic normalization.

By minimizing the edit distance to human-generated ground truth models, we utilize GED and RGED as proxies for structural fidelity and logical correctness. However, it is important to note that these metrics do not explicitly capture higher-level qualitative factors such as visual readability, layout simplicity, or cognitive load, which are emphasized in studies like those by Huang and Kumar \cite{huangNewQualityMetrics2009} and Pavlicek et al. \cite{pavlicekBusinessProcessModel2017}.

\subsection{Comparison with Baseline XML Editing}

To establish the effectiveness of BPMN Assistant, we conducted a comparative analysis between our atomic editing approach and a full-regeneration baseline (implemented here as direct XML generation). This baseline serves as a proxy for conversational refinement workflows used by state-of-the-art tools such as BPMN-Chatbot \cite{kopkeIntroducingBPMNChatbotEfficient2024} and ProMoAI \cite{kouraniProMoAIProcessModeling2024}. Both systems prompt the LLM to produce a revised structured representation of the process conditioned on the current model state (e.g., intermediate JSON in BPMN-Chatbot, or constrained code/POWL construction in ProMoAI). BPMN-Chatbot explicitly targets token efficiency via its intermediate JSON representation, whereas ProMoAI primarily leverages formal structure and constrained generation to improve reliability.

For the XML baseline, we utilized a system prompt instructing the model to act as a BPMN expert and output valid BPMN 2.0 XML directly, without intermediate steps or function calling. This represents the standard 'zero-shot' approach common in current LLM applications. For this comparison, we selected a representative set of diagram modification tasks and executed them using both representations. We evaluated the approaches using RGED and its derived similarity score, as defined in Equations~\eqref{eq:rged} and~\eqref{eq:similarity}.

\subsection{Generation Accuracy}

For the purposes of this evaluation, a "failure" is defined as any generation event that resulted in an unrenderable model. This encompasses both syntactic invalidity—such as unclosed XML tags or malformed JSON objects that prevent parsing—and structural inconsistencies, specifically reference hallucinations where sequence flows target node identifiers that do not exist within the element set. Any output falling into these categories was recorded as a failure in Tables~\ref{tab:evaluation} and \ref{tab:summary} and treated as a non-functional result.

The comparative analysis revealed that our JSON-based representation achieved an average similarity score of 0.72 compared to 0.70 for direct XML generation. While this indicates a slight numerical advantage for JSON, the difference is minimal, suggesting that both representations perform similarly in terms of structural accuracy. However, JSON demonstrated greater reliability, with fewer total failures across models.

\begin{table}[H]
\centering
\begin{tabular}{@{}lcccc@{}}
\toprule
\textbf{Model} & \textbf{JSON} & \textbf{XML} & \textbf{Failures (JSON)} & \textbf{Failures (XML)} \\
\midrule
GPT-5.1 & 0.73 & 0.70 & 0 & 0 \\
GPT-5 mini & 0.77 & 0.68 & 0 & 0 \\
GPT-4o & 0.71 & 0.68 & 0 & 0 \\
Claude 4.5 Sonnet & 0.80 & 0.76 & 0 & 0 \\
Claude 3.5 Sonnet & 0.72 & 0.72 & 0 & 0 \\
Gemini 2.0 Flash & 0.70 & 0.66 & 0 & 0 \\
Llama 3.3 70B Instruct & 0.68 & 0.70 & 0 & 4 \\
Qwen 2.5 72B Instruct & 0.69 & 0.72 & 2 & 6 \\
DeepSeek V3 & 0.72 & 0.69 & 0 & 1 \\
\bottomrule
\end{tabular}
\vspace{0.2cm}
\caption{Evaluation results showing similarity scores for different models using both our JSON-based BPMN representation and direct XML generation approaches. Each model was evaluated on 60 BPMN generation tasks.}
\label{tab:evaluation}
\end{table}

\begin{table}[h!]
\centering
\begin{tabular}{@{}lcc@{}}
\toprule
\textbf{Modality} & \textbf{Average Score} & \textbf{Total Failures} \\
\midrule
JSON & 0.72 & 2 \\
XML  & 0.70 & 11 \\
\bottomrule
\end{tabular}
\vspace{0.2cm}
\caption{Average similarity scores and total failures per modality across all generated models.}
\label{tab:summary}
\end{table}

Beyond similarity scores, JSON continues to outperform XML in terms of efficiency, a trend that holds even with newer, faster models. Latency measurements represent the total API response time recorded via the respective commercial providers (OpenAI, Anthropic, Google, and Fireworks AI). This metric reflects the real-world performance experienced by end-users of cloud-based LLM services. As shown in Table~\ref{tab:summary_stats}, JSON-based BPMN generation achieved a mean latency of 13.42 seconds compared to 24.82 seconds for XML. Additionally, while JSON requires a richer input prompt (averaging 2,678 tokens), it produces significantly more concise outputs (688 tokens vs 1,832 tokens). This reduction in output tokens is particularly advantageous for cost scaling, as output tokens are typically significantly more expensive than input tokens. Furthermore, the financial overhead of the increased input context is effectively mitigated by prompt caching technologies, which drastically reduce the cost of processing the static schema definitions and instructions required by our approach.

\begin{table}[H]
\centering
\begin{tabular}{@{}lrr@{}}
\toprule
\textbf{Metric} & \textbf{JSON} & \textbf{XML} \\
\midrule
Mean Latency (seconds) & 13.42 & 24.82 \\
\midrule
Average Input Tokens & 2678.63 & 474.05 \\
Average Output Tokens & 688.04 & 1832.48 \\
\bottomrule
\end{tabular}
\vspace{0.2cm}
\caption{Summary of latency and token usage between JSON and XML representations for BPMN generation.}
\label{tab:summary_stats}
\end{table}

\begin{figure}[ht]
\centering
\includegraphics[width=0.9\textwidth]{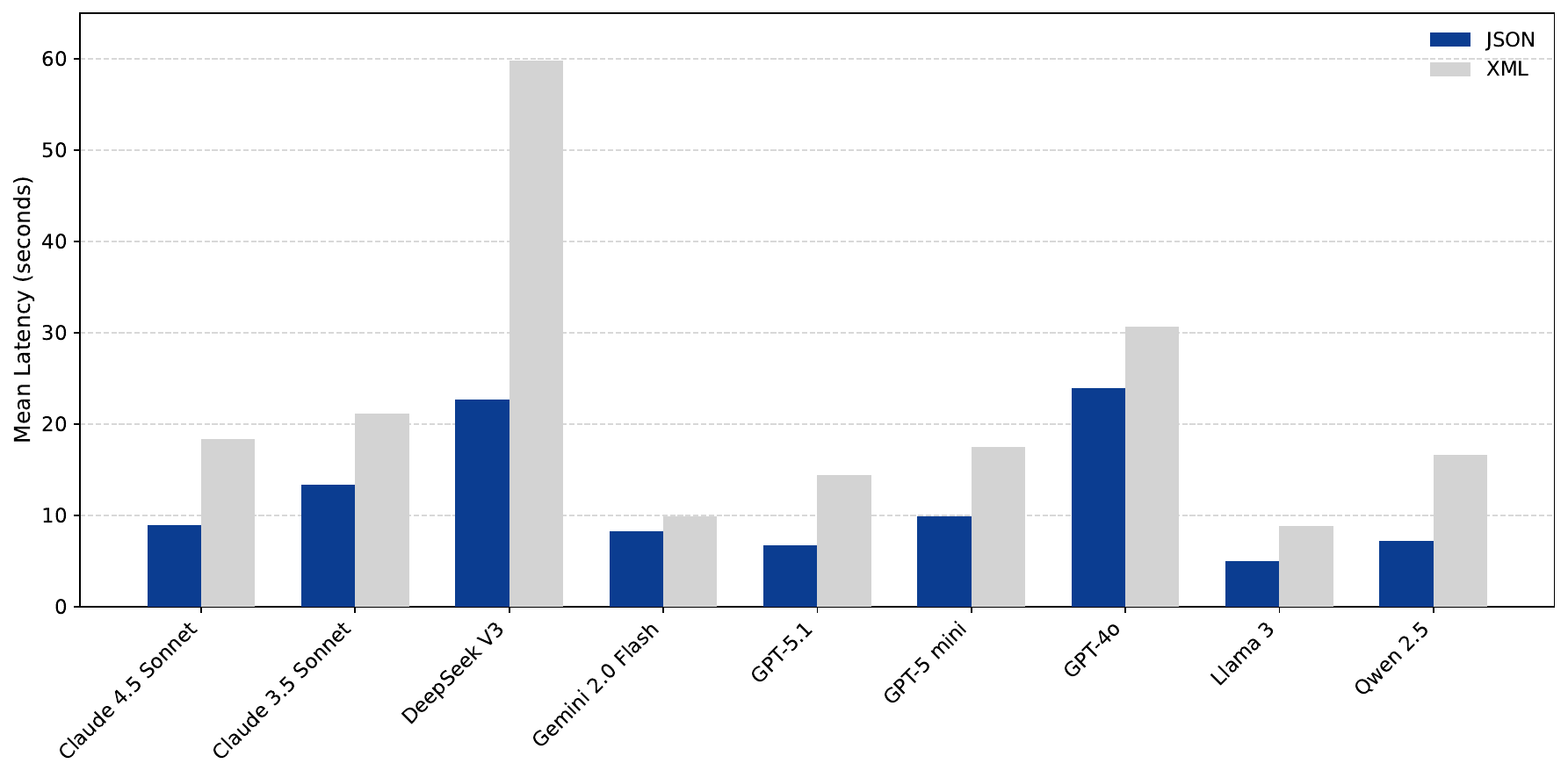}
\caption{Latency comparison between JSON and XML-based BPMN generation.}
\label{fig:latency_comparison}
\end{figure}

\begin{figure}[htbp]
\centering

\begin{subfigure}{\textwidth}
    \centering
    \includegraphics[width=0.9\linewidth]{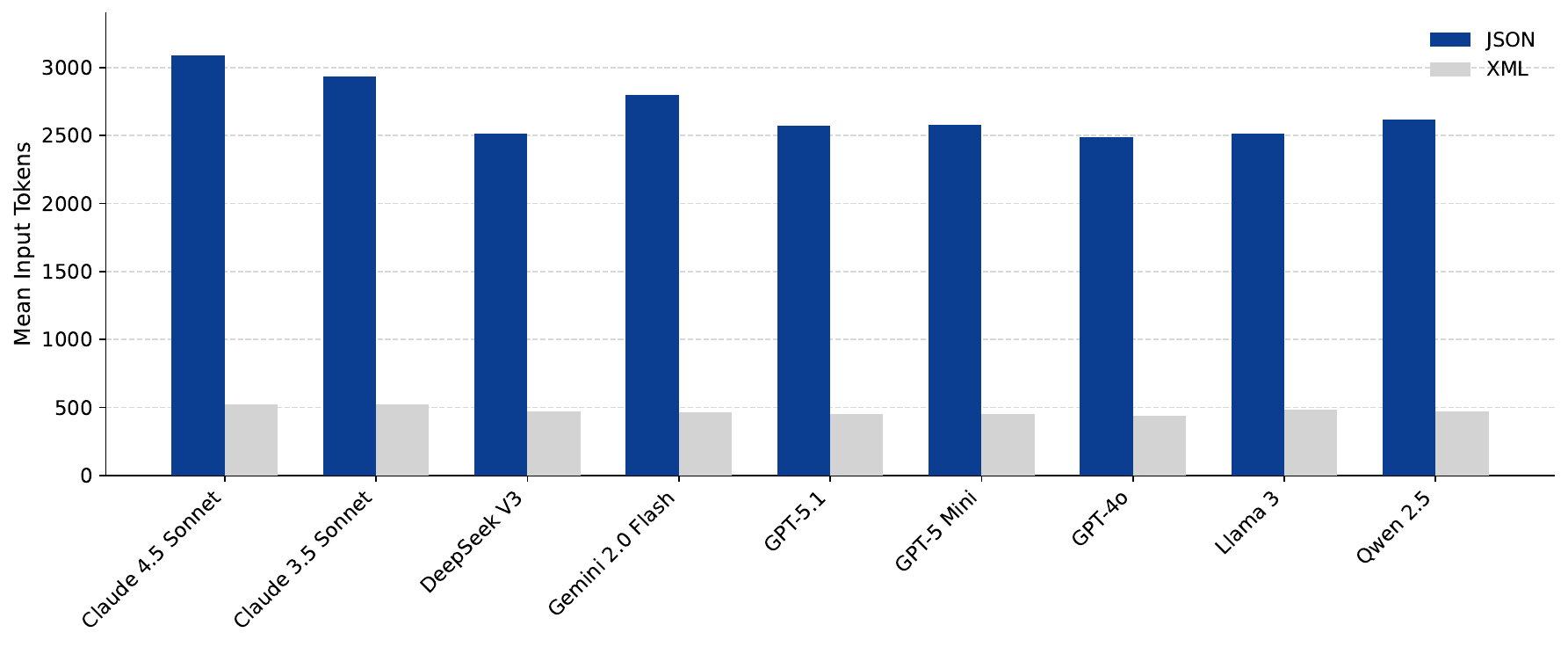}
    \caption{Comparison of mean input tokens.}
    \label{fig:token_usage_input}
\end{subfigure}

\vspace{1em}

\begin{subfigure}{\textwidth}
    \centering
    \includegraphics[width=0.9\linewidth]{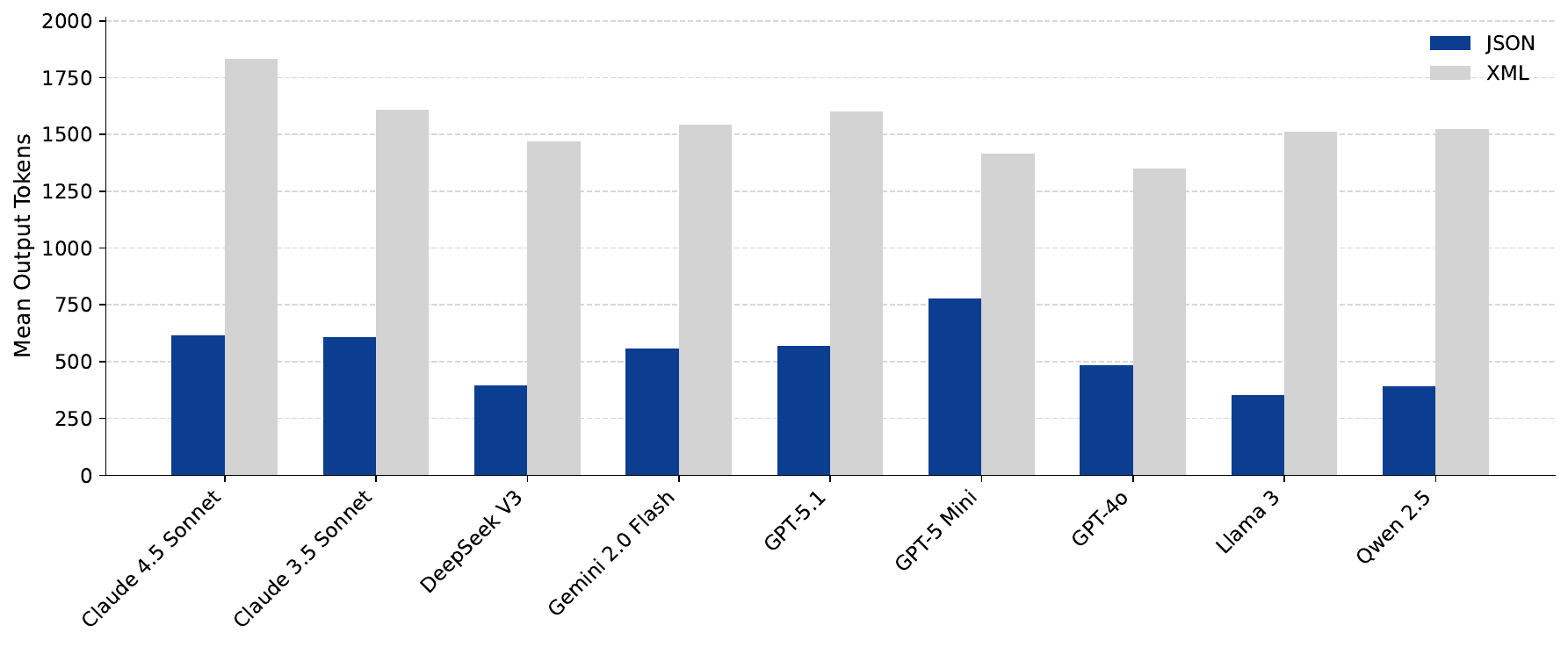}
    \caption{Comparison of mean output tokens.}
    \label{fig:token_usage_output}
\end{subfigure}

\caption{Token usage comparison for input and output tokens in JSON and XML-based BPMN generation. (a) shows the higher input token requirement for JSON prompts. (b) shows the significantly more concise output generated by the JSON approach.}
\label{fig:token_usage_comparison}
\end{figure}

\subsection{Editing Capabilities}
In addition to diagram generation, we evaluated the models' ability to interpret natural language editing instructions and perform the requested modifications. For this evaluation phase, we manually curated a dataset of 40 specific modification requests targeting various process elements. We employed a binary success/fail metric, as editing tasks involve correctly understanding the request and applying appropriate modifications to an existing diagram without corrupting the remaining structure.

To determine validity, we employed a two-stage verification process. First, a Python-based automated validator checked all outputs for syntactic correctness and referential integrity (e.g., ensuring no broken XML tags or references to non-existent IDs). Outputs failing this stage were automatically classified as failures. Second, the syntactically valid outputs underwent expert manual verification to assess semantic correctness. This step involved a binary check against the editing prompt (e.g., verifying that a "delete task" operation actually removed the target node and correctly reconnected the surrounding sequence flows). Given the objective nature of these boolean operations, a single expert evaluator was deemed sufficient.

The results, summarized in Table~\ref{tab:editing}, show that models consistently achieved higher success rates when using our JSON-based approach compared to direct XML manipulation. The performance gap is particularly evident in open-weight models; for instance, DeepSeek V3 achieved a 50\% success rate with JSON but failed almost entirely with XML (8\%). This suggests that the structured intermediate representation helps mitigate the complexities of verbose XML syntax, allowing models to focus on the logical changes rather than the syntactic overhead of the BPMN standard.

Even for flagship models like Claude 4.5 Sonnet, which achieved an 85\% success rate in both modalities, the JSON approach offers superior programmatic verifiability. Because the JSON output adheres to a strict internal schema, the system can validate logical consistency (e.g., ensuring all branch targets exist) before conversion. In contrast, validating generated XML often requires complex parsing that may fail on minor syntax hallucinations.

\begin{table}[h!]
\centering
\begin{tabular}{@{}lcccc@{}}
\toprule
\textbf{Model} & \textbf{JSON} & \textbf{XML} \\
\midrule
GPT-5.1 & 0.83 & 0.75 \\
GPT-5 mini & 0.73 & 0.53 \\
GPT-4o & 0.55 & 0.30 \\
Claude 4.5 Sonnet & 0.85 & 0.85 \\
Claude 3.5 Sonnet & 0.68 & 0.65 \\
Gemini 2.0 Flash & 0.45 & 0.33 \\
Llama 3.3 70B Instruct & 0.38 & 0.30 \\
Qwen 2.5 72B Instruct & 0.38 & 0.25 \\
DeepSeek V3 & 0.50 & 0.08 \\
\bottomrule
\end{tabular}
\vspace{0.2cm}
\caption{Success rates for diagram editing based on natural language instructions, comparing our JSON-based approach with direct XML editing. Each model was evaluated on 40 diverse editing tasks.}
\label{tab:editing}
\end{table}

\begin{figure}[ht]
\centering
\includegraphics[width=0.9\textwidth]{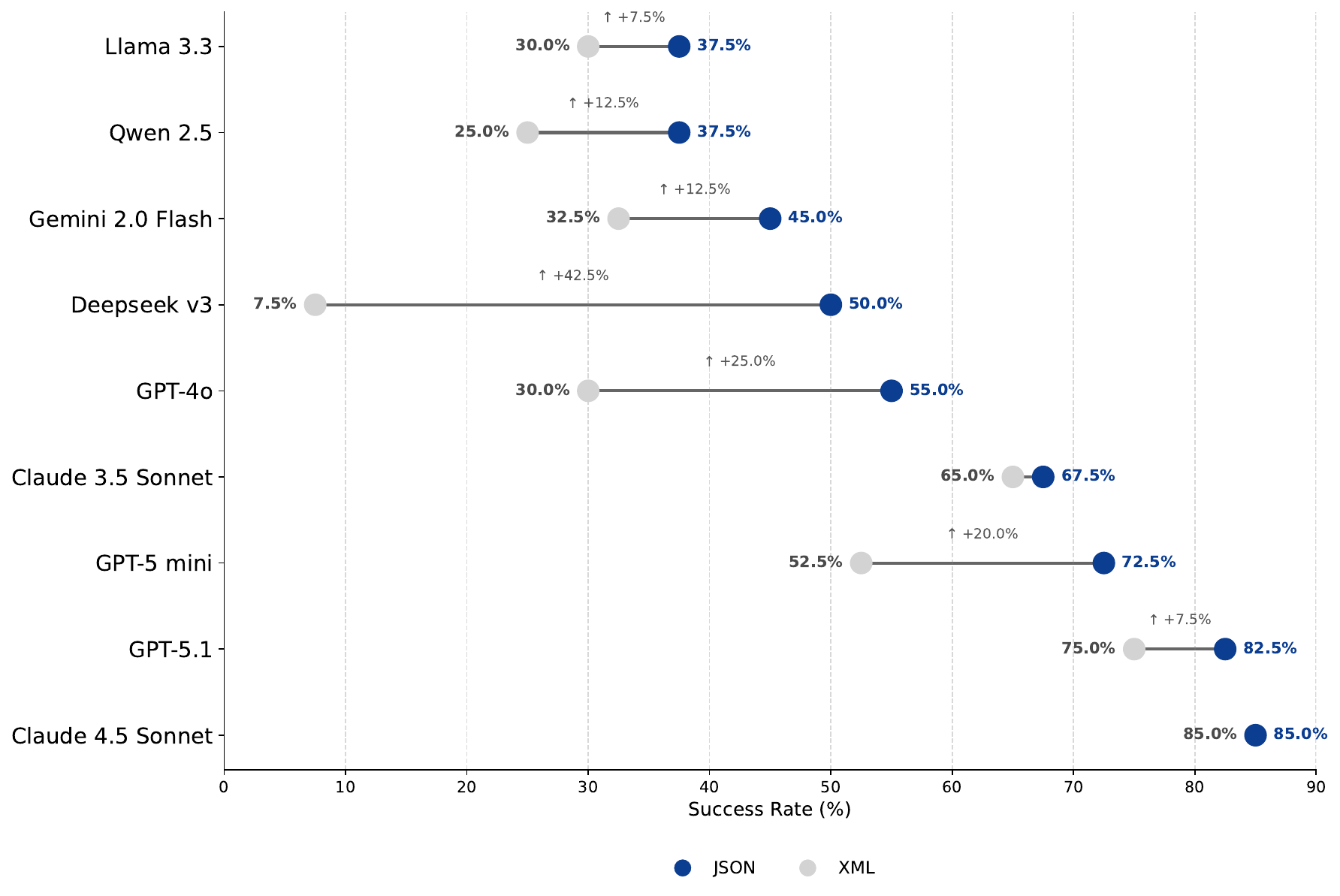}
\caption{Success rate comparison for diagram editing tasks between JSON and XML-based BPMN representations.}
\label{fig:edit_success_rate}
\end{figure}

As detailed in Table~\ref{tab:performance_comparison}, the trade-off for this reliability is context size. The JSON editing approach requires supplying the full BPMN intermediate representation and various process examples in the prompt, resulting in an average of 22,071 input tokens compared to 5,149 for XML. However, this 'up-front' cost pays dividends in speed: the JSON approach reduces generation latency by nearly 43\% (20.35s vs 35.63s) and reduces output verbosity by over 75\% (607 vs 2,630 tokens). In an interactive tool, this lower latency is critical for user experience.

\begin{table}[h!]
\centering
\begin{tabular}{@{}lrr@{}}
\toprule
\textbf{Metric} & \textbf{JSON} & \textbf{XML} \\
\midrule
Average Latency (s) & 20.35 & 35.63 \\
Average Input Tokens & 22,071.44 & 5,149.42 \\
Average Output Tokens & 607.92 & 2,630.44 \\
\bottomrule
\end{tabular}
\vspace{0.2cm}
\caption{Comparative performance metrics between JSON and XML-based approaches for BPMN editing.}
\label{tab:performance_comparison}
\end{table}

\section{Discussion}

These findings empirically validate the initial conceptual assumption of this work, namely that constraining LLM interaction through structured representations leads to more reliable and controllable BPMN model manipulation.

\subsection{The Efficacy of Structured Representations (RQ1)}
Our evaluation reveals a distinct dichotomy between generation and editing performance. In de novo generation tasks, the difference between JSON and XML approaches was marginal (e.g., GPT-5.1 achieved 0.73 with JSON vs 0.70 with XML). This suggests that for "tabula rasa" tasks, modern LLMs are sufficiently capable of managing XML syntax when generating from scratch.

However, the editing tasks expose the fragility of direct XML manipulation. The JSON-based approach achieved consistently higher success rates across all models. We attribute this difference to the verbosity and structural fragility of direct XML manipulation. When editing XML, an LLM must navigate verbose opening and closing tags, often losing track of the hierarchical structure or hallucinating invalid ID references. In contrast, our JSON schema and the associated atomic editing functions (e.g., \texttt{add\_element}) force the model to focus on the logical operation rather than the syntactic overhead. This structural constraint effectively acts as a guardrail, ensuring that modifications remain valid by design.

\subsection{Democratizing Process Modeling (RQ2)}
Perhaps the most significant finding is the impact of our approach on open-weights models. While frontier proprietary models like Claude 4.5 Sonnet performed admirably in XML editing (85\%), open-weights models such as DeepSeek V3 struggled significantly, achieving only an 8\% success rate with XML. However, when switching to our JSON-based function calling approach, DeepSeek V3's success rate jumped to 50\%.

This has profound implications for enterprise adoption. Many organizations in regulated industries (finance, healthcare) cannot send sensitive process data to external APIs like OpenAI or Anthropic due to data privacy concerns. Our results demonstrate that by using a structured intermediate representation, organizations can deploy locally hosted, open-weights models to perform complex process modeling tasks that were previously the domain of massive frontier models. This effectively lowers the barrier to entry for secure, on-premise AI process assistants.

\subsection{Efficiency and Latency Trade-offs (RQ3)}
The shift to a JSON-based editing workflow introduces a counter-intuitive trade-off: we drastically increase the input context size to decrease latency. As shown in Table~\ref{tab:performance_comparison}, the JSON approach requires supplying the entire current state of the process, leading to a $\sim 4\times$ increase in input tokens compared to the XML baseline.

However, in the current landscape of LLM economics and performance, this trade-off is highly favorable. Input tokens are computationally cheaper and faster to process than output tokens. By shifting the complexity to the input (the context) and restricting the output to concise JSON function calls, we achieved a 43\% reduction in total latency (20.35s vs 35.63s). For interactive tools where user experience depends on responsiveness, this reduction is critical. Furthermore, the 75\% reduction in output tokens leads to significant cost savings at scale, as output tokens typically cost 3-4 times more than input tokens across major providers. This efficiency is further amplified by prompt caching, which allows the static JSON schema definitions—the bulk of our input context—to be processed at a fraction of the standard input cost.

\subsection{Comparison with Existing Approaches}
Unlike previous tools such as BPMN-Chatbot~\cite{kopkeIntroducingBPMNChatbotEfficient2024} and ProMoAI~\cite{kouraniProMoAIProcessModeling2024}, which rely on regenerating the full process definition (via JSON or code) for every refinement, our system separates the editing logic from the model generation. While ProMoAI leverages formal methods (POWL) and BPMN-Chatbot utilizes intermediate JSON to optimize model creation, both approaches fundamentally treat modification as a conversational regeneration task. In contrast, BPMN Assistant implements a dedicated intermediate layer that supports atomic function calling, enabling targeted and deterministic updates without the overhead or instability associated with regenerating the entire process state for minor edits.

\section{Limitations and Future Work}

BPMN Assistant, while demonstrating promising results, has limitations that should be acknowledged. Although the system supports a robust set of elements, including intermediate events (timer, message) and inclusive gateways, the current implementation does not support collaboration diagrams (pools and lanes) or complex artifacts like data objects. This restricts the system's application in scenarios requiring multi-participant modeling. However, as the focus of this work is on executable process orchestration, this omission does not impact the primary research objectives.

Due to the lack of semantic evaluation, the results should be considered experimental, as there is no guarantee that the business logic will be correctly integrated into the generated models. This limitation is particularly relevant when assessing the practical applicability of the generated diagrams in real-world scenarios.

The performance of BPMN Assistant is heavily dependent on the underlying LLM. Our evaluation revealed significant variations in performance across different models, with Claude 4.5 Sonnet and GPT-5.1 showing particularly strong results. Further investigation into the specific capabilities and limitations of different LLMs in the context of process modeling would provide valuable insights for future improvements.

Human-computer interaction (HCI) studies would be necessary to properly evaluate the usability of the system for non-technical users. While BPMN Assistant aims to lower the barrier to process modeling through natural language interaction, comprehensive user studies would be required to validate this claim.

Another area of concern is the reliance on clear and unambiguous natural language input. Users might encounter difficulties in achieving this clarity, especially in multilingual or context-specific scenarios where language and terminology nuances can introduce ambiguity.

It is important to note that this work deliberately does not focus on semantic evaluation (e.g., using Petri nets) or conformance checking (e.g., process mining techniques), performance and execution time analysis (e.g., simulation studies), or comprehensive usability evaluation. The approach used in this research was selected considering the research objectives, the method of model generation, and existing evaluation methods cited in the literature. Future work could address these limitations by expanding the range of supported BPMN elements, incorporating semantic evaluation techniques, conducting usability studies, and optimizing the system for specific LLMs based on their strengths in process modeling tasks.

\section{Conclusion}

This paper presented BPMN Assistant, a novel approach to business process modeling that leverages LLMs to bridge the gap between natural language descriptions and formal BPMN representations. Through a comprehensive evaluation using state-of-the-art models, including GPT-5.1, Claude 4.5 Sonnet, and DeepSeek V3, we demonstrated that using a simplified JSON structure instead of raw XML yields significant benefits in reliability and efficiency.

Our results indicate that while direct XML generation is feasible for creating simple models from scratch, it is often insufficient for iterative modification tasks. The proposed JSON-based approach outperformed or matched direct XML manipulation across all tested models in editing scenarios. While frontier models like Claude 4.5 Sonnet achieved parity in both modalities, the JSON advantage was critical for other models. This was most pronounced in open-weights models, where the structured approach enabled DeepSeek V3 to achieve a viable 50\% success rate compared to a near-total failure (8\%) with standard XML prompting. This suggests that intermediate representations are a key enabler for deploying cost-effective, locally hosted models in process automation workflows.

In terms of performance, the trade-off between increased input context and reduced output complexity proved highly advantageous. By shifting the computational burden to the input context, our approach reduced editing latency by approximately 43\% and decreased output token usage by over 75\%. These efficiency gains are critical for the practical adoption of LLM-based tools in interactive, real-time environments.

Looking ahead, future work will focus on conducting comprehensive usability studies to validate the system's effectiveness with non-technical business users. Additionally, we aim to further optimize the JSON schema to better handle complex cyclic dependencies and improve the system's robustness when handling highly ambiguous natural language descriptions.

\newpage

\printbibliography

\newpage

\appendix
\section{Appendix: Process Example}
\label{sec:appendix-process-example}

The following examples demonstrate how the JSON intermediate representation handles complex BPMN constructs, including cyclic flows (loops) and event definitions.

\subsection*{Example 1: Cyclic Flow (Loops)}
This example demonstrates the use of the \texttt{next} attribute to model a loop where a student must retake an exam upon failure. This illustrates the system's ability to handle non-hierarchical flows.

\textbf{Textual description:} \\
"Someone starts a process by entering an exam room. After that, they take the test. Once the exam is finished, their score is checked. If they scored more than 50\%, their grade is recorded and the process ends. But if they fail, they have to go back to the beginning and take the exam again."

\noindent
\textbf{JSON Representation:}
\begin{verbatim}
{
    "process": [
      {
        "type": "startEvent",
        "id": "start"
      },
      {
        "type": "task",
        "id": "task1",
        "label": "Enter exam room"
      },
      {
        "type": "task",
        "id": "task2",
        "label": "Take test"
      },
      {
        "type": "exclusiveGateway",
        "id": "eg",
        "label": "Score more than 50%?",
        "has_join": false,
        "branches": [
          {
            "condition": "Yes",
            "path": [
              {
                "type": "task",
                "id": "task3",
                "label": "Enter grade"
              },
              {
                "type": "endEvent",
                "id": "end1"
              }
            ]
          },
          {
            "condition": "No",
            "path": [],
            "next": "task1" // Redirects flow back to the first task
          }
        ]
      }
    ]
}
\end{verbatim}

\subsection*{Example 2: Event-Driven Workflow}
This example demonstrates the support for specific event definitions (Timer and Message events) required for executable process models.

\textbf{Textual description:} \\
"A batch process starts at midnight using a timer. The system processes pending orders and waits for 5 minutes. Then it sends a completion notification and waits to receive a confirmation message before ending."

\noindent
\textbf{JSON Representation:}
\begin{verbatim}
{
  "process": [
    {
      "type": "startEvent",
      "id": "start",
      "label": "Start at midnight",
      "eventDefinition": "timerEventDefinition"
    },
    {
      "type": "serviceTask",
      "id": "task1",
      "label": "Process pending orders"
    },
    {
      "type": "intermediateCatchEvent",
      "id": "timer1",
      "label": "Wait 5 minutes",
      "eventDefinition": "timerEventDefinition"
    },
    {
      "type": "intermediateThrowEvent",
      "id": "message1",
      "label": "Send completion notification",
      "eventDefinition": "messageEventDefinition"
    },
    {
      "type": "intermediateCatchEvent",
      "id": "message2",
      "label": "Wait for confirmation",
      "eventDefinition": "messageEventDefinition"
    },
    {
      "type": "endEvent",
      "id": "end"
    }
  ]
}
\end{verbatim}

\end{document}